\pdfoutput=1

\documentclass[11pt]{article}

\usepackage[final]{acl}

\usepackage{times}
\usepackage{latexsym}

\usepackage[T1]{fontenc}

\usepackage[utf8]{inputenc}

\usepackage{microtype}

\usepackage{inconsolata}

\usepackage{graphicx}

\usepackage{arydshln}
\usepackage{booktabs}
\usepackage{multirow}
\usepackage{multicol}
\usepackage{amsmath}
\usepackage{amssymb}
\newcommand{\probP}{\text{I\kern-0.15em P}}

\title{JNLP at SemEval-2025 Task 11: Cross-Lingual Multi-Label Emotion Detection Using Generative Models}

\author{
 Jieying Xue\textsuperscript{1} \and
 Phuong Minh Nguyen\textsuperscript{2,1} \and
 Minh Le Nguyen\textsuperscript{1} \and
 Xin Liu\textsuperscript{3}
\\
\\
 \textsuperscript{1}Japan Advanced Institute of Science and Technology\\
 \textsuperscript{2}ROIS-DS Center for Juris-Informatics, NII, Tokyo, Japan\\
 \textsuperscript{3}National Institute of Advanced Industrial Science and Technology
\\
 \texttt{\{xuejieying,phuongnm,nguyenml\}@jaist.ac.jp}, \texttt{xin.liu@aist.go.jp}   \\
 \small{
   \textbf{Correspondence:} \href{mailto:phuongnm@jaist.ac.jp}{phuongnm@jaist.ac.jp}
 }
}

\begin{document}
\maketitle
\begin{abstract}
With the rapid advancement of global digitalization, users from different countries increasingly rely on social media for information exchange. In this context, multilingual multi-label emotion detection has emerged as a critical research area.
This study addresses SemEval-2025 Task 11: Bridging the Gap in Text-Based Emotion Detection. Our paper focuses on two sub-tracks of this task: (1) Track A: Multi-label emotion detection, and (2) Track B: Emotion intensity.
To tackle multilingual challenges, we leverage pre-trained multilingual models and focus on two architectures: (1) a fine-tuned BERT-based classification model and (2) an instruction-tuned generative LLM. Additionally, we propose two methods for handling multi-label classification: the \textit{base} method, which maps an input directly to all its corresponding emotion labels, and the \textit{pairwise} method, which models the relationship between the input text and each emotion category individually.
Experimental results demonstrate the strong generalization ability of our approach in multilingual emotion recognition. In Track A, our method achieved Top 4 performance across 10 languages, ranking 1st in Hindi. In Track B, our approach also secured Top 5 performance in 7 languages, highlighting its simplicity and effectiveness\footnote{Our code is available at \url{https://github.com/yingjie7/mlingual_multilabel_emo_detection}}.

\end{abstract}
\section{Introduction}
With the rapid proliferation of social media, particularly in the context of global digital communication, online platforms have emerged as the primary medium for information dissemination \citep{nandwani2021review}. Users from diverse linguistic backgrounds frequently express their opinions through comments, highlighting the growing need for cross-lingual sentiment detection \citep{nandwani2021review}. Consequently, multilingual sentence-level sentiment analysis has become a critical task for tracking public sentiment \citep{wankhade2022survey}.
Sentiment analysis is one of the most extensively studied applications in natural language processing (NLP). In text emotion recognition, it is common for a single sentence to express multiple emotions with varying intensities \citep{deng2020multi}. However, developing reliable multi-label emotion analysis systems remains particularly challenging due to the scarcity of training data, especially for low-resource languages. Additionally, pre-trained language models often have limited knowledge of these languages, further complicating the task.
To address these challenges, this paper presents our approach for SemEval-2025 Task 11, "Bridging the Gap in Multilingual Multi-Label Emotion Detection from Text Using Large Language Models" \cite{muhammad-etal-2025-semeval}. 
We participated in two tracks: Track 1 (Multi-Label Emotion Detection) using the BRIGHTER dataset, which includes 28 languages \citep{muhammad2025brighterbridginggaphumanannotated,belay-etal-2025-evaluating}, and Track 2 (Emotion Intensity Prediction), which covers 11 languages.

In this study, to address the challenges of multilingual sentiment analysis, we leveraged pre-trained models (such as RoBERTa) and large language models (LLMs) to perform multi-label sentiment analysis on both high-resource languages like English and Chinese, as well as low-resource languages such as African languages.  We formulated multi-label emotion recognition as a text generation task. To overcome the challenge of limited training data, we utilize the capabilities of multilingual pre-trained language models to enhance both semantic understanding and the recognition of emotional tone, especially in low-resource languages. Furthermore, to tackle the multi-label classification challenge, we propose two methods: the \textit{pairwise} method and the \textit{base} method. Our findings also indicate that training the model on a combined multilingual dataset improves performance compared to training on individual language datasets. 
We present experiments comparing the applicability of these methods and conduct ablation studies to validate their effectiveness. Our approach demonstrates strong performance in both multi-label emotion recognition and emotion intensity detection. In Track A, it achieved top-four rankings in 10 languages, including first place in Hindi. In Track B, our method ranked within the top five for 7 languages, further highlighting its simplicity and effectiveness. Moreover, our approach exhibits strong generalization across both competition sub-tasks, making it particularly beneficial for low-resource languages.

\section{Background}

Sentence-level sentiment analysis (SLSA) has advanced significantly with the rise of deep learning and multilingual sentiment detection.
Early research primarily focused on extracting handcrafted sentiment features such as n-grams \citep{tripathy2016classification}, lexicons, rule-based heuristics \citep{chikersal2015sentu} to enhance SVM-based classifiers \citep{kumari2017sentiment} and deep neural networks, such as CNNs and RNNs 
\citep{chikersal2015sentu, minaee2019deep}. 
Nonetheless, their reliance on static word embeddings limited their ability to handle complex linguistic phenomena such as long-range dependencies and cross-lingual variations.
To address these limitations, researchers turned to Transformer-based pretrained language models (PLMs) such as BERT \citep{devlin-etal-2019-bert} and T5 \citep{raffel2020exploring}, which more effectively capture fine-grained emotional representations \citep{zhou2016attention, li2018hierarchical} by modeling richer linguistic semantics.
In multilingual sentiment analysis, models like mT5 \citep{xue2020mt5} and XLM-RoBERTa (XLM-R) \citep{conneau-etal-2020-unsupervised} further advanced the field by learning cross-lingual representations, making them the standard for multilingual applications \citep{hu2020xtreme}.
More recently, the widespread adoption of large language models (LLMs) such as LLaMA 2 \citep{touvron2023llama} has driven major breakthroughs in various NLP tasks \citep{upadhye2024sentiment, Sharma2023MiningTF}. These models exhibit remarkable zero-shot and few-shot learning capabilities, making them highly adaptable to new sentiment analysis tasks \citep{10.1145/3639233.3639353}. 
Furthermore, in the domain of Emotion Recognition in Conversations, LLMs have been leveraged with prompt-based techniques to extract latent supplementary knowledge from text, injecting this information to facilitate emotion recognition \cite{xue2024bioserc}.
In the broader NLP landscape, various methodologies including fine-tuning, prompting, transfer learning, and domain adaptation have been pivotal in adapting pre-trained LLMs for sentiment analysis across specific domains and languages.  

However, as most PLMs are predominantly pretrained on English text, their effectiveness in multilingual sentiment analysis is often limited without additional fine-tuning is performed to optimize performance across diverse linguistic contexts \citep{zhang2023don}. Numerous studies have explored leveraging embeddings from LLMs for sentiment classification, using various low-resource datasets to assess their adaptability across languages \citep{dadure-etal-2025-sentiment, 10.1145/3605889}. 
In our work, we leverage BERT-based multilingual models to extend multi-label classification tasks, enabling knowledge transfer across languages. By integrating LLMs, we also present a \textit{pairwise emotional recognition} method, which efficiently captures both emotional intensity and sentiment polarity within each sentence. This approach ensures that the model concentrates on one label at a time. Additionally, we reformulate the multi-label classification task as a text generation problem, enhancing the model’s adaptability and generalization across NLP tasks.

\section{System Description}
In this work, the target task involves the perception of emotions in various languages, which aims to identify the emotion that most people would attribute to the speaker based on a given sentence or short text snippet. Given a text input ($x$), a machine learning system needs to retrieve all the multi-label emotions ($y_e$) expressed in the given text (Track A) and the intensity ($y_{i}$) of each class (Track B). 
\subsection{System Overview}

In general, we leverage the capabilities of pre-trained multilingual models to tackle cross-lingual challenges. Our system primarily focuses on two architectures: fine-tuning BERT-based classification models \citep{devlin-etal-2019-bert} and instruction fine-tuning generative LLMs, building upon recent SOTA methods in the field of emotion recognition \citep{xue2024bioserc}. To handle multi-label classification, we design two strategies: (1) the \textit{base} method, which maps a given input to all its corresponding labels, and (2) the \textit{pairwise} method, which models the relationship between the input text and each label class individually.
\begin{equation} \resizebox{0.8\hsize}{!}{$
\textrm{\textit{base}:}  \quad \probP_{A}(\{y_e\} \mid x) \,\qquad\probP_{B}(\{\langle y_e, y_i\rangle\} \mid x)  
$}
\end{equation}\vspace{-20pt}
\begin{equation}\resizebox{0.88\hsize}{!}{$ 
\textrm{\textit{pairwise}\vspace{5pt}:} \quad  \probP_{A}(\{0, 1\} \mid x, y_e)\quad  \probP_{B}(y_i \mid x, y_e)
$}
\end{equation} 
where $x$ is the given input text;  $\probP_{A}, \probP_{B}$ are probability models for track A and B; $y_e, y_i$ are the emotional label and its corresponding emotional intensity from a pre-defined label set, respectively. 
\subsection{Methods}
\paragraph{BERT-based method.} For the baseline, we design a BERT-based multi-label classification model. In detail, fully connected layers with nonlinear activation  functions (sigmoid ($\sigma$) and tanh) are added to the top layer of the BERT architecture to transform the \texttt{[CLS]} feature vector \citep{devlin-etal-2019-bert} from the hidden representation to the output dimension (number of labels). 
\begin{align}
     h^{CLS}, h^{words} &= \mathrm{BERT}( x ) \\
     h^{out} &=  \sigma (\textrm{tanh} ( h^{CLS} \cdot W^{h}) \cdot W^{o})  \label{eq:u_i}
\end{align}
Finally, during the fine-tuning process, the learnable weights ($W^*$) are optimized using cross-entropy loss on annotated data to maximize the log-likelihood of the model.

\paragraph{LLM-based method.} Leveraging the robust natural language understanding capabilities of large language models (LLMs) \citep{touvron2023llama}, we employ instruction prompting (highlighted in blue in Table~\ref{tab:ft_prompting}) to guide the model in comprehending the task requirements. Our methodology follows instruction fine-tuning as outlined by \citet{flant5_instruction}, using a \textit{causal language modeling} objective to train the LLM to generate emotional label text, which is highlighted in red in Table~\ref{tab:ft_prompting}.
\begin{align}
s &=\textrm{instruction-prompting}(x, y) \\
\probP(s)&=\Pi_{z=1}^{|s|} \probP(s_z|s_0, s_1, ..., s_{z-1})
\end{align}
where $s, x$ represent a sequence of tokens, and $z$ denotes the token index within the prompting input (Table~\ref{tab:ft_prompting}).
To optimize efficiency, we employ LoRA \citep{hu2022lora}, a lightweight training methodology that reduces the number of trainable parameters. The fine-tuned LLM is designed to learn the distribution of emotional labels (or emotional intensity) based on the given prompt ($s$). During inference, the emotional label ($y$), which is omitted from the input prompt, is generated by the fine-tuned model. 
\begin{table}[htbp]
\small
\centering
\resizebox{\columnwidth}{!}{
\begin{tabular}{|p{1.15\columnwidth}|}
    \hline
    \textit{system}\\
    \textcolor{blue}{You are an expert in analyzing the emotions expressed in a natural sentence. The emotional label set includes \{anger, fear, joy, sadness, surprise\}. Each sentence may have one or more emotional labels, or none at all.} \\  
    \textit{user}\\
    Given the sentence: ``\texttt{\{input text: $x$\}}'', which emotions are expressed in it?\\
    \textit{assistant} \\
    \textcolor{red}{\texttt{\{emotional label in text: $y_e$ or  $\langle y_e, y_i\rangle$\}}}\\
    \hline
\end{tabular}
}
\caption{{Instruction prompting with the \textit{base} template (track A).}\label{tab:ft_prompting}}
\end{table}

\begin{table}[!htbp]
\small
    \centering
    \resizebox{\columnwidth}{!}{%
        \begin{tabular}{lcllp{0.43\columnwidth}lll}
            \hline  \multirow{1}{*}{\textbf{Task}}&\textbf{Strategy}&\textbf{Input} &\textbf{Output} &\textbf{Output example} \\
            \hline
            Track A & base &$x$ & $\{y_e\}$ &  \textit{``disgust, sadness''} \\
            Track B & base & $x$ &$\{\langle y_e,  y_i\rangle\} $&\textit{``moderate degree of anger, low degree of sadness''} \\
            Track A & pairwise &$x,\,  y_e$& $\{0, 1\}$ & \textit{``yes''}  \\ 
            Track B & pairwise &$x,\, y_e$& $y_i$ & \textit{``moderate''}\\
            \hline 
        \end{tabular} 
        } 
        \caption{{Examples of output format for text generation.}\label{tab_text_output}}
\end{table}

As outlined in the overview section, we have devised two approaches, \textit{base} and \textit{pairwise}, to tackle this task. Both approaches employ the same training techniques across tracks A and B and between the two approaches themselves. We provide example outputs designed for both tracks in Table~\ref{tab_text_output}.
Detailed examples for each track are provided in Appendix~\ref{appendix_example}.

\section{Experimental Setup}
\paragraph{Dataset.} To evaluate our methods, we use the original emotional data provided by the SemEval Task 11 organization. This dataset consists of three subsets: training, development, and test sets, spanning two competition phases: development and test. However, to ensure greater generalization, we consistently set aside 10\% of the training data from each language as an internal development set. This held-out portion is used for hyper-parameter tuning, ensuring that the optimized checkpoints are selected based on this internal dev set. Additionally, to handle multilingual data, we design two settings: (1) \textit{separated langs}, where a separate model is trained for each language, and (2) \textit{mixed langs}, where a single model is trained to learn all languages simultaneously.




\begin{table*}[htbp]
\small
    \centering
    \resizebox{\textwidth}{!}{%
        \begin{tabular}{lrrrrrrrrrrrrrrrrrrrrrrrrrr}
        
            \hline  \multirow{1}{*}{\textbf{Model}}&\textbf{Strategy}&\textbf{Data}&\textbf{afr}&\textbf{amh}&\textbf{arq}&\textbf{ary}&\textbf{chn}&\textbf{deu}&\textbf{eng}&\textbf{esp}&\textbf{hau}&\textbf{hin}&\textbf{ibo}&\textbf{kin}&\textbf{mar}&\textbf{orm}&\qquad\\
            \hline
            \textit{(development)} \\
            Qwen 32b  & pairwise & separated langs  &\textcolor{red}{0.5143}&0.5049&\textcolor{red}{0.6574}&0.5242&0.6909&0.7187&\textcolor{red}{0.8189}&0.8366&0.5724&0.8694&0.5049&0.4274&0.9507 &-\\
            Qwen 32b& base & separated langs& 0.4610 &-&-&-&-&-&0.8054&-&-&-&-&-&-&-&\\
            Qwen 32b & base & mixed langs &0.5140&0.557&0.64&0.537&\textcolor{red}{0.732}&0.677&0.751&\textcolor{red}{0.839}&0.57&\textcolor{red}{0.899}&\textcolor{red}{0.509}&\textcolor{red}{0.477}&\textcolor{red}{0.959}&0.478\\
            Qwen 14b  & base &mixed langs &0.4320&0.594&0.588&\textcolor{red}{0.567}&0.643&0.691&0.743&0.835&0.606&0.887&0.498&0.454&0.924&0.503\\
            xml-roberta & base &mixed langs &0.5070&\textcolor{red}{0.66}&0.607&0.548&0.623&0.654&0.703&0.786&\textcolor{red}{0.687}&0.855&0.488&0.328&0.948&\textcolor{red}{0.513}\\
            \hline
            JNLP \textit{(test)}& && 0.5925 & 0.6767 & 0.6407 & 0.609 & 0.6805 & 0.6990$^*$ & 0.8036 & 0.8303 & 0.6504 & \textit{0.9257} & 0.5404 & 0.4289 & 0.878 & 0.573 \\\hline 
        \end{tabular} 
        } 
        \\[2pt]
    \resizebox{\textwidth}{!}{%
        \begin{tabular}{lrrrrrrrrrrrrrrrrrrrrrrrrrr}
        
            \hline  \multirow{1}{*}{\textbf{Model}}& \textbf{Strategy}&\textbf{Data}&\textbf{pcm}&\textbf{ptbr}&\textbf{ptmz}&\textbf{ron}&\textbf{rus}&\textbf{som}&\textbf{sun}&\textbf{swa}&\textbf{swe}&\textbf{tat}&\textbf{tir}&\textbf{ukr}&\textbf{vmw}&\textbf{yor}&\textbf{Average}\\
            \hline
            \textit{(development)} \\
            Qwen 32b& pairwise & separated langs&0.6202&\textcolor{red}{0.6407}&0.5161&\textcolor{red}{0.7548}&0.8809&0.3571&0.5307&0.2658&\textcolor{red}{0.5915}&0.6282&0.4581&0.6761&0.1265&\textcolor{red}{0.4554}&\textcolor{red}{0.5933} \\
            Qwen 32b& base & separated langs& 0.611 &-&-&0.7230&-&-&-&-&-&-&-&-&-&-&-\\
            Qwen 32b & base & mixed langs  &\textcolor{red}{0.638}&0.546&\textcolor{red}{0.571}&-&\textcolor{red}{0.902}&0.416&\textcolor{red}{0.557}&0.332&0.509&0.72&0.429&0.639&0.114&0.355& \textcolor{red}{0.5933}\\
            Qwen 14b & base & mixed langs  &0.622&0.576&0.553&-&0.895&0.394&0.51&0.319&0.494&\textcolor{red}{0.764}&0.485&0.64&\textcolor{red}{0.19}&0.348&0.5863\\
            xml-roberta  &base & mixed langs & 0.574&0.502&0.579&-&0.876&\textcolor{red}{0.499}&0.539&\textcolor{red}{0.348}&0.501&0.692&\textcolor{red}{0.5}&0.594&0.074&0.198&0.5718\\
            \hline
            JNLP \textit{(test)}& && 0.6343 & 0.6184 & 0.4535 & 0.7787 & 0.8912 & 0.4965 & 0.4596 & 0.2949 & 0.6186 & 0.7223 & 0.4849 & 0.6873$^*$ & 0.2261 & 0.3608 & 0.6163 \\
            \hline
        \end{tabular}
        } 
    \caption{
    Results of Sub-task A.   For a fair comparison, the \textit{average} column is computed based on all languages except for \textit{orm, ron, ptbr}, and \textit{ptmz}, as these languages are missing in some settings. The red color indicates the best setting used for submission to obtain the test result. The notation (-) indicates that the experiment was not conducted. The asterisk ($^*$) denotes results obtained during the post-evaluation phase.  \label{tab:all_resultsA} }
\end{table*} 

\begin{table*}[htbp]
\small
    \centering
    \resizebox{\textwidth}{!}{%
        \begin{tabular}{lrrrrrrrrrrrrrrrrrrrrrrrrrr}
        
            \hline  \multirow{1}{*}{\textbf{Model}}&\textbf{Strategy}&\textbf{Data}&\textbf{amh}&\textbf{arq}&\textbf{chn}&\textbf{deu}&\textbf{eng}&\textbf{esp}&\textbf{hau}&\textbf{ptbr}&\textbf{ron}&\textbf{rus}&\textbf{ukr}&\textbf{Average}\\
            \hline
            \textit{(development)} \\
            Llama2-13b & pairwise &separated langs & - & 0.4411 & \textcolor{red}{0.73857} & 0.6197 & \textcolor{red}{0.8207} & 0.7221 & 0.5691 & 0.4938 & 0.691 & 0.8719 & 0.6229 & 0.6757 \\ 
            Qwen-32b & pairwise  & separated langs &0.5433  & 0.6147 & 0.75 & 0.6793 & 0.8101  & 0.7715 & 0.6143 & - & \textcolor{red}{0.7245} & 0.9051 & 0.6428 & 0.7234 \\
            Qwen-32b &base & mixed langs & 0.542&0.566&0.711&0.658&0.802&0.761&0.595&\textcolor{red}{0.718}&-&0.898&0.659& 0.7063\\
            Qwen-32b &pairwise & mixed langs &  \textcolor{red}{0.563} & \textcolor{red}{0.627} & 0.727 & \textcolor{red}{0.705} & 0.787 & \textcolor{red}{0.779} & \textcolor{red}{0.665} & 0.6 & - & \textcolor{red}{0.906} & \textcolor{red}{0.694} & \textcolor{red}{0.7363} \\
            \hline
            JNLP \textit{(test)}&&& 0.6038 & 0.5873 & 0.6589 & 0.725 & 0.8129 & 0.7747 & 0.6496 & 0.6512 & 0.7055 & 0.9074 & 0.6719 & 0.7044\\
            \hline
        \end{tabular}  
        } 
    \caption{
    Results of Sub-task B. The meanings of the denotations and colors are the same as in Table~\ref{tab:all_resultsA}.  \label{tab:all_resultsB} }
\end{table*} 
\paragraph{Evaluation Metric.} According to the competition guidelines, the evaluation metric for Track A is the macro-averaged F1-score, while for Track B, it is the Pearson correlation coefficient between the predicted and gold-standard labels.
\paragraph{Experimental Environments.}  We implement all our experiments using widely adopted libraries such as PyTorch and HuggingFace. For pretrained LLMs, we primarily experiment with XLM-RoBERTa-Large, Llama2 (7B-13B), and Qwen2.5 (14B–32B). For hyper-parameters, we train the model with a learning rate of $3e^{-4}$, using the AdamW optimization algorithm, 5–6 epochs. 

\section{Results}
Overall, we evaluate our methods and their variants on the development set to select the best model and setting for each language (indicated in red in Tables~\ref{tab:all_resultsA},~\ref{tab:all_resultsB}) for the final test submission.
\subsection{Track A: Multi-label Emotion Detection.}

\paragraph{Development Result.} As shown in Table~\ref{tab:all_resultsA}, we conducted experiments using both the \textit{base} and \textit{pairwise} methods on Qwen-32B, Qwen-14B, and RoBERTa models. The results indicate that the Qwen models with the \textit{pairwise} method achieved the best overall performance. However, in datasets where the majority of samples contain zero or only one emotion label, the \textit{base} method outperformed the \textit{pairwise} approach. We attribute this to the fact that the \textit{pairwise} method is inherently more suited for multi-label emotion recognition tasks.
Additionally, in low-resource languages, LLMs performed poorly, whereas the RoBERTa-based approach yielded better results.
 
\paragraph{Test Result.} Overall, our approach achieved 4th place in Track A for CHN, ESP, PCM, and PT-BR, secured 3rd place in ARQ, ARY, RON, and RUS, and ranked 2nd and 1st for SWE and HIN, respectively. These results demonstrate the strong generalization ability of our method, highlighting its simplicity and efficiency.

\subsection{Track B: Emotion Intensity.}
The results of Track A demonstrated the strength of LLMs compared to the XLM-RoBERTa model, leading us to primarily experiment with LLMs rather than XLM-RoBERTa in this track. 

\textit{Development Result.} As shown in Table~\ref{tab:all_resultsB}, we conducted experiments using both the \textit{base} and \textit{pairwise} strategies on LLaMA 2 and Qwen-32B models. The results indicate that Qwen-32B outperforms LLaMA 2, and the \textit{pairwise} strategy consistently achieves better overall performance compared to the \textit{base} method. 

\textit{Test Result.}  In Track B across 11 languages, our model achieved 3rd place in Ukrainian (ukr) and Algerian Arabic (arq), 4th place in Romanian (ron), and 5th place in Russian (rus), Brazilian Portuguese (ptbr), English (eng), and German (deu). With top-five rankings in seven languages, these results demonstrate the effectiveness and generalizability of our approach.
\subsection{Result Analyses}
\paragraph{Strategies Comparison.}
    
\begin{figure}
        \centering 
        \includegraphics[width=0.45\linewidth, keepaspectratio, 
                trim={0.8cm 0 0cm 0}, page=1, clip=true]{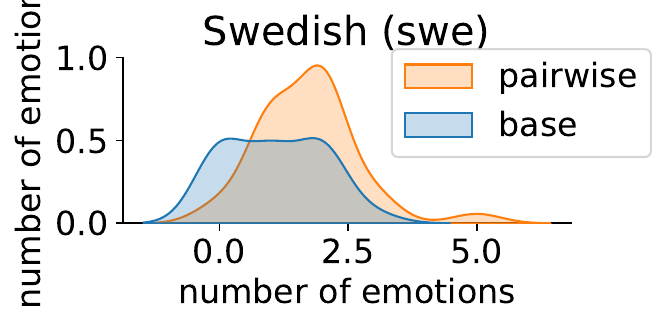}
        \includegraphics[width=0.45\linewidth, keepaspectratio, 
                trim={0.8cm 0 0cm 0}, page=1, clip=true]{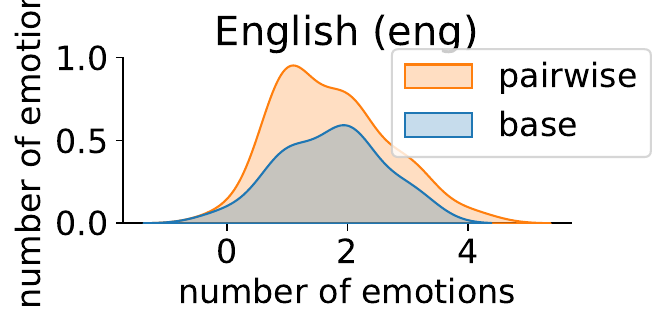} 
        \includegraphics[width=0.45\linewidth, keepaspectratio, 
                trim={0.8cm 0 0cm 0}, page=1, clip=true]{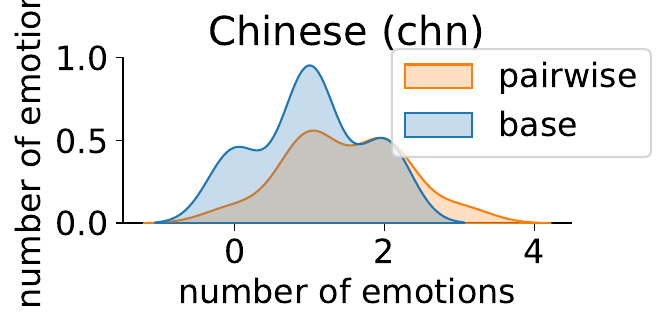}
        \includegraphics[width=0.45\linewidth, keepaspectratio, 
                trim={0.8cm 0 0cm 0}, page=1, clip=true]{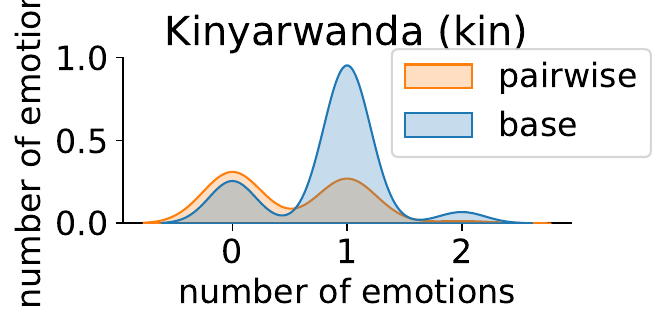}
        \caption{Distribution of improved samples between strategies \textit{base} and \textit{pairwise} with respect to the number of emotions (track A).} 
        \label{fig_improved_distribution_A}
    \end{figure}

\begin{figure}
        \centering 
        \includegraphics[width=0.45\linewidth, keepaspectratio, 
                trim={0.8cm 0 0cm 0}, page=1, clip=true]{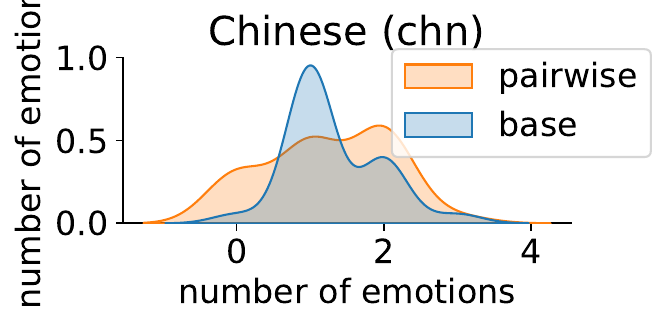}
        \includegraphics[width=0.45\linewidth, keepaspectratio, 
                trim={0.8cm 0 0cm 0}, page=1, clip=true]{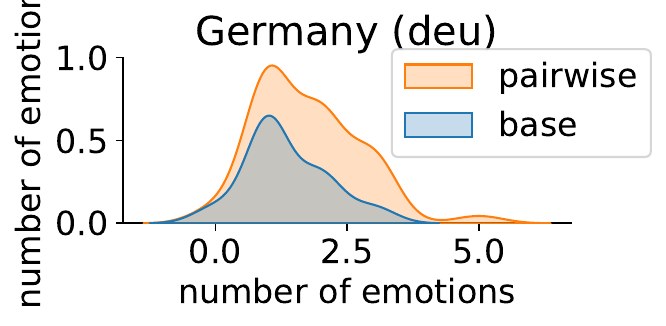} 
        \includegraphics[width=0.43\linewidth, keepaspectratio, 
                trim={0.8cm 0 0cm 0}, page=1, clip=true]{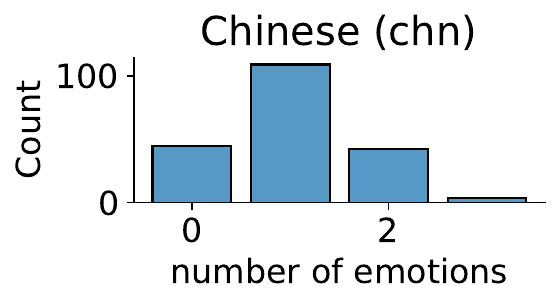}
        \includegraphics[width=0.43\linewidth, keepaspectratio, 
                trim={0.8cm 0 0cm 0}, page=1, clip=true]{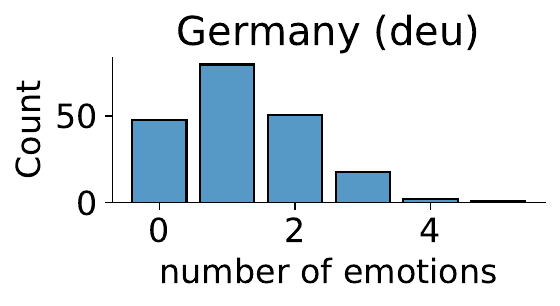} 
        \caption{Distribution of improved samples between strategies \textit{base} and \textit{pairwise} with respect to the number of emotions (track B).} 
        \label{fig_improved_distribution_B}
    \end{figure}

To gain a comprehensive understanding of the base and pairwise strategies, we conducted experiments to analyze the distribution of improved examples, measured by the F1 score for each sample, across four languages: English, Swedish, Chinese, and Kinyarwanda (Figure~\ref{fig_improved_distribution_A} for Track A and Figure~\ref{fig_improved_distribution_B} for Track B). Our findings indicate that the pairwise strategy predominantly improves samples in languages that convey various emotions within a sentence, particularly in English and Swedish. Conversely, in languages or datasets with a limited variety of emotional labels, such as Chinese and Kinyarwanda (where each sample typically contains 0 to 2 emotions), the base strategy demonstrates a distinct advantage.
We argue that this is because the pairwise strategy evaluates only one emotion at a time, making it more sensitive to label imbalance, which in turn leads to lower performance in languages with a limited variety of emotion labels compared to the base strategy. In contrast, the base strategy generates all emotions present in a sample simultaneously, highlighting its advantage in languages with fewer distinct emotion categories.

    \begin{figure}
        \centering 
        \includegraphics[width=0.7\linewidth, keepaspectratio, 
                trim={0.8cm 0 0cm 0}, page=1, clip=true]{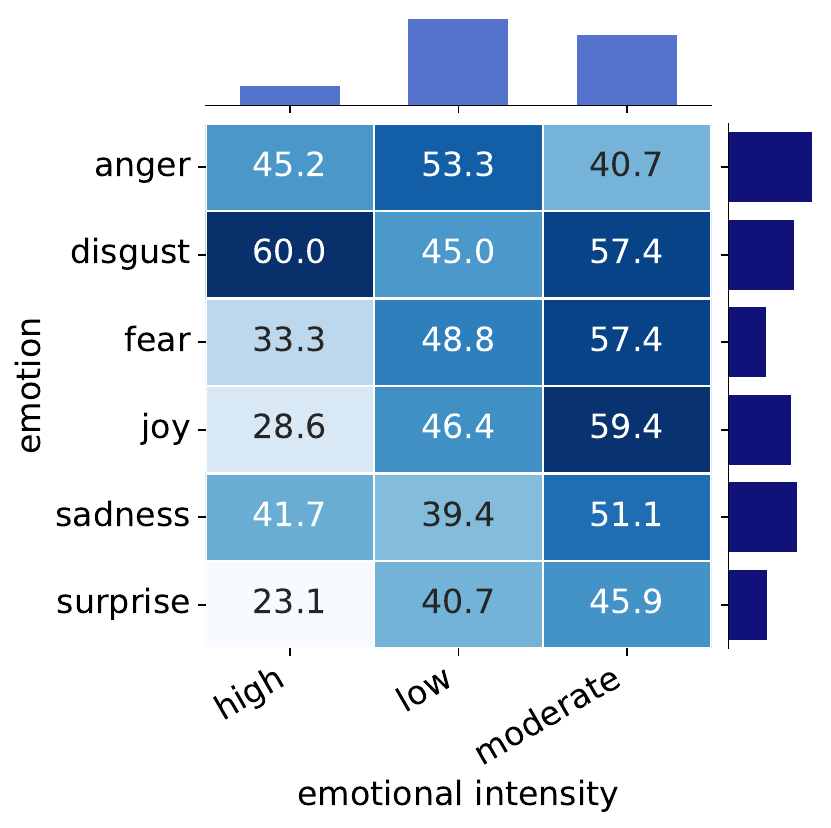}
        \caption{Overall performance of the pairwise strategy across all emotional labels in all languages (Track B).} 
        \label{fig_dist_emotions}
    \end{figure}

\paragraph{Emotional Type.}
To evaluate the model's effectiveness concerning emotional intensity across various emotions, we plotted the distribution of emotional labels alongside their corresponding intensities (Fig~\ref{fig_dist_emotions}). We conducted experiments by aggregating all the languages and examined the correlation between performance and emotions, as well as their respective intensities. Our analysis revealed that classes with limited data, such as high-surprise or high-joy, typically exhibited poorer performance in our system. Conversely, the major emotional class, \textit{``disgust''}, achieved the highest performance compared to other emotional classes, such as surprise, particularly high-surprise.
\paragraph{Mixed languages.} In both Sub-tasks A and B, mixed-language training, where a single model is fine-tuned for all languages, demonstrates superior performance compared to training separate models (Tables~\ref{tab:all_resultsA},~\ref{tab:all_resultsB}). This improvement can be attributed to a more balanced distribution of emotion types across languages and the model’s enhanced ability to generalize across linguistic variations.

\section{Conclusion}
In this work, we present a multilingual emotion recognition system for SemEval-2025 Task 11, which demonstrates strong performance and remains competitive with the top-performing teams. 
To address multilingual challenges, we design two architectures, BERT-based and LLM-based, and introduce two strategies, \textit{pairwise} and \textit{base}, for handling the multi-label classification task. 
We conduct extensive experiments to analyze the effectiveness and limitations of each approach, aiming to provide valuable insights for multilingual emotion recognition research. 
The results validate the simplicity and effectiveness of our methods, highlighting their strong generalization ability and applicability to other tasks.
\paragraph{Limitations and potential improvements.} 
Despite the promising results achieved in our work, several limitations remain.
First, the LLM-based approach is heavily dependent on the knowledge and capabilities within the LLMs themselves, which may limit its adaptability to evolving data. Furthermore, compared to BERT-based methods, LLM-based approach incurs higher computational costs, making it less efficient for large-scale or real-time applications.

There are several directions for future improvement.
One potential enhancement lies in the utilization of finer-grained information contained in the logits output. Specifically, for the LLM Pairwise strategy, instead of relying solely on the final "yes" or "no" response, it is better to aggregate the logits corresponding to the tokens generating "yes" and compute a probability distribution via softmax. This would enable a more nuanced and probabilistic interpretation of the model’s predictions, potentially improving robustness.

Another limitation is the incomplete handling of label imbalance. Our current framework does not fully address the issue, which may cause the model to overfit to dominant emotional categories. Future work could incorporate targeted data augmentation strategies, such as generating additional samples for underrepresented emotions, to mitigate this imbalance and enhance the overall performance and stability of the system.




\bibliography{custom}

\appendix
\section{Examples of prompting\label{appendix_example}} 

\begin{table}[htbp]
\small
\centering
\resizebox{\columnwidth}{!}{
\begin{tabular}{|p{1.\columnwidth}|}
    \hline
    \textit{system}\\
    {You are an expert in analyzing the emotions expressed in a natural sentence. The emotional label set includes \{anger, fear, joy, sadness, surprise\}. Each sentence may have one or more emotional labels, or none at all.} \\  
    \textit{user}\\
    Given the sentence: ``\texttt{bro dont do this to us}'', which emotions and their corresponding intensities are expressed in it?\\
    \textit{assistant} \\
    {\texttt{fear}}\\
    \hline
\end{tabular}
}\\[10pt]
\resizebox{\columnwidth}{!}{
\begin{tabular}{|p{1.\columnwidth}|}
    \hline
    \textit{system}\\
    {You are an expert in analyzing the emotions expressed in a natural sentence. The emotional label set includes \{anger, fear, joy, sadness, surprise\}. Each sentence may have one or more emotional labels, or none at all.} \\  
    \textit{user}\\
    Given the sentence: ``\texttt{I could not unbend my knees.}'', is the emotion anger expressed in it?\\
    \textit{assistant} \\
    {\texttt{No}}\\
    \hline
\end{tabular}
}
\caption{{Instruction prompting template in track A of  \textit{base} (top) and \textit{pairwise} (bottom) strategies, respectively.}\label{tab_example_A_base}}
\end{table} 
\begin{table}[htbp]
\small
\centering
\resizebox{\columnwidth}{!}{
\begin{tabular}{|p{1.\columnwidth}|}
    \hline
    \textit{system}\\
    {You are an expert in analyzing the emotions expressed in a natural sentence. The emotional label set includes \{anger, fear, joy, sadness, surprise\}, with three levels of intensity: low, moderate, and high. Each sentence may have one or more emotional labels, or none at all.} \\  
    \textit{user}\\
    Given the sentence: ``\texttt{A penny hit me square in the face.}'', which emotions and their corresponding intensities are expressed in it?\\
    \textit{assistant} \\
    {\texttt{moderate degree of anger, low degree of sadness}}\\
    \hline
\end{tabular}
}\\[10pt]
\resizebox{\columnwidth}{!}{
\begin{tabular}{|p{1.\columnwidth}|}
    \hline
    \textit{system}\\
    {You are an expert in analyzing the emotions expressed in a natural sentence. The emotional label set includes \{anger, fear, joy, sadness, surprise\}, with three levels of intensity: low, moderate, and high. Each sentence may have one or more emotional labels, or none at all.} \\  
    \textit{user}\\
    Given the sentence: ``\texttt{Totally creeped me out.}'', what is the intensity of the emotion fear expressed in it?\\
    \textit{assistant} \\
    {\texttt{high}}\\
    \hline
\end{tabular}
}
\caption{{Instruction prompting template in track B of  \textit{base} (top) and \textit{pairwise} (bottom) strategies, respectively.}\label{tab_example_B_base}}
\end{table} 

\end{document}